\documentclass[twoside]{article}
\usepackage[accepted]{aistats2015arxiv} %

\usepackage{amsmath,amssymb,amsthm}

\usepackage{microtype}

\usepackage[utf8]{inputenc} 
\usepackage{graphicx} %
\usepackage{subfigure}
\newlength{\sfigwidth}
\usepackage{algorithm,algorithmicx}
\usepackage[noend]{algpseudocode} %
\newcommand{\algcomment}[1]{\Comment{\textit{#1}}}
\algnewcommand{\LineComment}[1]{\(\triangleright\) \textit{#1}}

\usepackage[authoryear,round]{natbib}
\usepackage{afterpage}
\usepackage[colorlinks,citecolor=blue,urlcolor=blue,linkcolor=blue]{hyperref}

\usepackage{wrapfig}
\usepackage{grffile}
\usepackage{float}
\usepackage{rotating}
\usepackage{enumitem} 
\usepackage{bm}
\usepackage{dsfont}
\usepackage{cite}
\usepackage{color}
\usepackage{soul}
\usepackage{wasysym}
\usepackage{url}
\usepackage{cases}

\usepackage[capitalize]{cleveref}  
\usepackage{mathtools}
\usepackage{amsthm}
\usepackage{thmtools}

\usepackage[bf,small]{caption} %
\DeclareCaptionType{copyrightbox}

\newcommand{\real}{\mathrm{I\kern-0.175em R}}
\def\T{\mathcal{T}}
\def\U{\mathcal{U}}

\def\bv{\bm{v}}

\def\bx{x}

\def\bz{\bm{z}}

\def\bX{X}

\def\Normal{\mathcal{N}}

\def\jdash{j^{\prime}}

\def\indicator{\mathds{1}}

\def\Reals{\mathbb{R}}
\renewcommand{\bx}{\bm{x}}
\renewcommand{\bX}{\bm{X}}

\newcommand{\bY}{{Y}}
\newcommand{\PT}{\mathsf{T}}
\newcommand{\TS}{\mathcal{T}}

\renewcommand{\T}{\mathsf{T}}

\newcommand{\leaf}[1]{\mathsf{leaves}(#1)}
\newcommand{\nonleaf}[1]{#1 \setminus \leaf{#1} }
\newcommand{\leafx}[1]{\mathsf{leaf}_{\TS}(#1)}

\newcommand{\depth}{\textsf{depth}}
\newcommand{\nparts}{C}
\newcommand{\cgm}{CGM}%
\newcommand{\growprune}{GrowPrune}

\newcommand{\mbart}{m}      %
\newcommand{\allj}{1:\mbart}
\newcommand{\param}{\mu}
\newcommand{\bparam}{\bm{\param}}
\newcommand{\ii}{(i)}
\newcommand{\iiold}{(i-1)}
\newcommand{\node}{\eta}
\newcommand{\ensemble}{\{\TS_j, \bparam_j\}_{j=1}^\mbart}
\newcommand{\ensemblenotj}{\{\TS_{\jdash}, \bparam_{\jdash}\}_{\jdash\neq j}}

\newcommand{\sdim}{\kappa}
\newcommand{\sloc}{\tau}
\newcommand{\bsdim}{\bm{\sdim}}
\newcommand{\bsloc}{\bm{\sloc}}

\def\wt{\tilde{w}}
\def\wtbar{\overline{\wt}}
\def\given{\,|\,}
\def\defas{:=}
\def\grow{\emph{grow}}
\def\prune{\emph{prune}}
\def\swap{\emph{swap}}
\def\change{\emph{change}}
\newcommand{\Cc}{c}
\newcommand{\cc}{{(c)}}
\newcommand{\ccone}{{(1)}}
\newcommand{\Ccd}{{c'}}
\newcommand{\ccd}{{(c')}}
\newcommand{\smcitervar}{t}
\newcommand{\smciter}{(\smcitervar)}
\newcommand{\smciterold}{(\smcitervar-1)}
\renewcommand\wt{w}
\def\[#1\]{\begin{align}#1\end{align}}

\def\reducevspace{\vspace{-0.1in}}

\def\ctslices{\emph{CTslices}}
\def\houses{\emph{CaliforniaHouses}}
\def\msd{\emph{YearPredictionMSD}}

\begin{document}

\twocolumn[
\aistatstitle{Particle Gibbs for Bayesian Additive Regression Trees}
\aistatsauthor{ Balaji Lakshminarayanan \And Daniel M. Roy\ \And Yee Whye Teh }
\aistatsaddress{ Gatsby Unit\\University College London \And Department of Statistical Sciences\\ University of Toronto \And Department of Statistics\\ University of Oxford } 
]

\begin{abstract} 
Additive regression trees are flexible non-parametric models %
and popular off-the-shelf tools for real-world non-linear regression. 
In application domains, such as bioinformatics, where there is also demand for probabilistic predictions with measures of uncertainty, 
the Bayesian additive regression trees (BART) model, introduced by \citet{Chipman10}, is increasingly popular.
As data sets have grown in size, however, the standard Metropolis--Hastings algorithms used to perform inference in BART are proving inadequate.
In particular, these Markov chains make local changes to the trees and suffer from slow mixing when the data are high-dimensional or the best-fitting trees are more than a few layers deep.
We present a novel sampler for BART based on the Particle Gibbs (PG) algorithm \citep{AndDouHol2010a}
and a top-down particle filtering algorithm for Bayesian decision trees \citep{LRT2013}.
Rather than making local changes to individual trees, the PG sampler proposes a complete tree to fit the residual. 
Experiments show that the PG sampler %
 outperforms existing samplers in many settings.
\end{abstract} 

\let\thefootnote\relax\footnote{%
This document is identical in content to, and should be cited as:
B.~Lakshminarayanan, D.~M.~Roy, and Y.~W.~Teh, 
\emph{Particle Gibbs for Bayesian Additive Regression Trees}, 
in Proceedings of the $18^{th}$ International Conference on Artificial Intelligence and Statistics (AISTATS) 2015, San Diego, CA, USA\@. JMLR: W\&CP volume 38. %
}

\reducevspace %
\reducevspace %
\reducevspace %
\reducevspace
\section{Introduction}
Ensembles of regression trees
are at the heart of many state-of-the-art approaches for nonparametric regression \citep{caruana2006empirical}, 
and 
can be broadly classified into two families: \emph{randomized independent regression trees}, wherein the trees are grown independently and predictions are averaged to reduce variance, and \emph{additive regression trees}, wherein each tree fits the residual not explained by the remainder of the trees. 
In the former category are bagged decision trees \citep{breiman1996bagging}, random forests \citep{breiman2001random}, extremely randomized trees \citep{geurts2006extremely}, and many others, while additive regression trees can be further categorized into those that are fit in a serial fashion, like gradient boosted regression trees \citep{friedman2001greedy}, and those fit in an iterative fashion, like Bayesian additive regression trees (BART) \citep{Chipman10} and additive groves~\citep{sorokina2007additive}. 

Among additive approaches, BART is extremely popular and has been successfully applied to a wide variety of problems including protein-DNA binding, credit risk modeling, automatic phishing/spam detection, and drug discovery \citep{Chipman10}. %
Additive regression trees must be regularized to avoid overfitting \citep{friedman2002stochastic}: in BART,
over-fitting is controlled by a prior distribution preferring simpler tree structures and non-extreme predictions at leaves.
At the same time, the posterior distribution underlying BART delivers a variety of inferential quantities beyond predictions, including credible intervals for those predictions as well as a measures of variable importance.
At the same time, BART has been shown to achieve predictive performance comparable to random forests, boosted regression trees, support vector machines, and neural networks~\citep{Chipman10}. 

The standard inference algorithm for BART is an iterative Bayesian backfitting Markov Chain Monte Carlo (MCMC) algorithm~\citep{hastie2000bayesian}. 
In particular, the MCMC algorithm introduced by \citet{Chipman10} proposes local changes to individual trees. 
This sampler 
can be computationally expensive for large datasets, and so recent work on scaling BART to large datasets \citep{pratola2013parallel} considers using only a subset of the moves proposed by \citet{Chipman10}. However, this smaller collection of moves has been observed to lead to poor mixing~\citep{pratola2013efficient} which in turn produces an inaccurate approximation to the posterior distribution. 
While a poorly mixing Markov chain might produce a reasonable prediction in terms of mean squared  error, BART is often used in scenarios where its users rely on posterior quantities, 
and so there is a need for \emph{computationally efficient samplers that mix well} across a range of hyper-parameter settings. 

In this work, we describe a novel sampler for BART based on (1) the Particle Gibbs (PG) framework proposed by \citet{AndDouHol2010a}
and (2) the top-down sequential Monte Carlo algorithm for Bayesian decision trees proposed by \citet{LRT2013}.
Loosely speaking, 
PG is the \emph{particle version of the Gibbs sampler}  where proposals from the exact conditional distributions are replaced by conditional versions of a sequential Monte Carlo (SMC) algorithm. %
The complete sampler follows the Bayesian backfitting MCMC framework for BART proposed by \citet{Chipman10}; the key difference is that trees are sampled using PG instead of the local proposals used by \citet{Chipman10}. 
Our sampler, which we refer to as \emph{PG-BART}, approximately samples complete trees from the conditional distribution over a tree fitting the residual.
As the experiments bear out, the PG-BART sampler explores the posterior distribution more efficiently than samplers based on local moves.
Of course, one could easily consider non-local moves in a Metropolis--Hastings (MH) scheme by proposing complete trees from the tree prior, however these moves would be rejected, leading to slow mixing, in high-dimensional and large data settings.
The PG-BART sampler succeeds not only because non-local moves are considered, but because those non-local moves have high posterior probability.
Another advantage of the PG sampler is that it only requires one to be able to sample from the prior and does not require evaluation of tree prior in the acceptance ratio unlike (local) MH\footnote{The tree prior term cancels out in the MH acceptance ratio if complete trees are sampled. However, sampling complete trees from the tree prior  would lead to very low acceptance rates as discussed earlier.}---hence PG can be computationally efficient in situations where the tree prior is expensive (or impossible) to compute, but relatively easier to sample from. 

The paper is organized as follows: in section \ref{sec:model}, we review the BART model; in section \ref{sec:inference}, we review the MCMC framework proposed by \citet{Chipman10} and describe the PG sampler in detail. %
 In section \ref{sec:experiments}, we present experiments that compare the %
  PG sampler to existing samplers for BART. 

\section{Model and notation}\label{sec:model}
In this section, we briefly review decision trees and the BART model. %
We refer the reader to the paper of \citet{Chipman10} for further details about the model. Our notation closely follows their's, but also borrows from \citet{LRT2013}.%

\subsection{Problem setup}
We assume that the training data consist of $N$ i.i.d.~samples $\bX=\{\bx_n\}_{n=1}^N$, where $\bx_n\in\Reals^D$, along with corresponding labels $\bY=\{y_n\}_{n=1}^N$, where $y_n\in\Reals$. We focus only on the regression task in this paper, although the PG sampler can also be used for classification by building on the work of \citet{Chipman10} and \citet{zhang2010bayesian}.

\newcommand{\LatentTree}{\mathcal T}

\subsection{Decision tree}
For our purposes, a decision tree is a hierarchical binary partitioning of the input space with axis-aligned splits.
The structure of the decision tree 
is 
a finite, rooted, strictly binary tree $\T$, i.e., a finite collection of nodes such that  1) every node $\node$ has exactly one \emph{parent} node, except for a distinguished \emph{root} node $\epsilon$ which has no parent, and 2) every node $\node$ 
is the parent of exactly zero or two \emph{children} nodes,
called the left child $\node0$ and the right child $\node1$.
Denote the leaves of $\T$ (those nodes without children) by $\leaf\T$.  Each node of the tree $\node\in\T$  is associated with a block $B_{\node}\subset\Reals^D$ of the input space as follows:  At the root, we have $B_{\epsilon}=\Reals^D$, while each \emph{internal} node $\node\in\nonleaf\T$ with two children represents a \emph{split} of its parent's block into two halves, with $\sdim_{\node}\in\{1,\dotsc,D\}$ denoting the dimension of the split, and $\sloc_{\node}$ denoting the location of the split.  In particular,
\begin{align}
B_{\node0} &= B_{\node}\cap\{\bz \in\Reals^D: z_{\sdim_{\node}}\le \sloc_{\node}\} \text{ and }\nonumber \\
B_{\node1} &= B_{\node}\cap\{\bz \in\Reals^D: z_{\sdim_{\node}}> \sloc_{\node}\}.
\end{align}
We call the tuple $\LatentTree=(\T,\bsdim,\bsloc)$ a \emph{decision tree}.  
   Note that the blocks associated with the leaves of the tree form a partition of $\Reals^D$.

A decision tree used for regression is referred to as a \emph{regression tree}. In a regression tree, each leaf node $\node\in\leaf{\PT}$ is associated with a real-valued parameter $\param_{\node}\in\Reals$.  Let $\bparam=\{\param_{\node}\}_{\node \in\leaf{\PT}}$ denote the collection of all parameters. Given a tree $\TS$ and a data point $\bx$, let 
$\leafx{\bx}$ be the unique leaf node %
$\eta \in\leaf{\PT}$ such that $\bx\in B_{\eta}$, and let $g( \,\cdot\,; \TS, \bparam)$ be the response function associated with $\TS$ and $\bparam$, given by
\[
g(\bx; \TS,\bparam)\defas\param_{\leafx{\bx}}.
\]

\subsection{Likelihood specification for BART}

BART is a %
 \emph{sum-of-trees} model, i.e., BART assumes that the label $y$ for an input $\bx$ is generated by an additive combination of $\mbart$ regression trees. More precisely,   
\begin{align}
y = \sum_{j=1}^{\mbart} g(\bx; \TS_j, \bparam_j) + e,
\end{align}
where $e \sim \Normal(0,\sigma^2)$ is an independent Gaussian noise term with zero mean and variance $\sigma^2$. 
Hence, the likelihood for a  training instance is
\begin{align*}
  \ell(y|\ensemble, \sigma^2, \bx) =  \Normal\bigl(y|\sum_{j=1}^{\mbart} g(\bx; \TS_j, \bparam_j), \sigma^2\bigr),
\end{align*}
and the likelihood for the entire training dataset is
\begin{align*}
    \ell(\bY|\ensemble, \sigma^2, \bX) = \prod_n \ell(y_n|\ensemble, \sigma^2, \bx_n).
\end{align*}

\newcommand{\splitprob}[2]{\mathit{s}\textrm{Pr}(#1, \TS, #2)}
\subsection{Prior specification for BART}\label{sec:prior specification}

The parameters of the BART model are the noise variance $\sigma^2$ and the regression trees $(\TS_j,\bparam_j)$ for $j = 1,\dotsc,m$.  The conditional independencies in the prior are captured by the factorization
\begin{align*}
    p(\ensemble, \sigma^2|\bX) = p(\sigma^2)\prod_{j=1}^{\mbart} p(\bparam_j|\TS_j) p(\TS_j|\bX). 
\end{align*}
The prior over decision trees $p(\TS_j=\{\T_j,\kappa_j,\tau_j\}|\bX)$ can be  described by the following generative process \citep{Chipman10,LRT2013}: 
Starting with a tree comprised only of a root node $\epsilon$, 
the tree is grown by deciding once for every node $\node$ whether to 1) \emph{stop} and make $\node$ a leaf, or 2) \emph{split}, making $\node$ an internal node, and add $\node0$ and $\node1$ as children.  The same stop/split decision is made for the children, and their children, and so on.  Let $\rho_\node$ be a binary indicator variable for the event that $\node$ is split. Then every node $\node$ is split independently with probability 
\[\label{eq:splitprob} %
p(\rho_\node=1)
 = &
      \frac{\alpha_s}{ (1+\depth(\node) )^{\beta_s}}
    \nonumber \\
    & \times \indicator[\text{valid split exists below $\node$ in $\bX$}],
\]
where the indicator $\indicator[...]$ forces the probability to be zero when every possible split of $\node$ is invalid, i.e., one of the children nodes contains no training data.\footnote{Note that $p(\rho_\node=1)$ depends on $\bX$ and the split dimensions and locations at the ancestors of $\node$ in $\TS$ due to the indicator function for valid splits. We elide this dependence to keep the notation simple.}
Informally, the hyperparameters $\alpha_s\in(0,1)$ 
and $\beta_s\in[0,\infty)$ control the depth and number of nodes in the tree. Higher values of $\alpha_s$ lead to deeper trees while higher values of $\beta_s$ lead to shallower trees.

In the event that a node $\node$ is split, the dimension $\sdim_{\node}$ and location $\sloc_{\node}$ of the split are assumed to be drawn independently from a uniform distribution over the set of all valid splits of $\node$.  
The
 decision tree prior is thus
\begin{align} %
p(\TS|\bX)=&
 \prod_{\node\in\nonleaf{\PT}}
    p(\rho_\node=1)
      \,\U(\sdim_{\node}) \, \U(\sloc_{\node}|\sdim_{\node})
   \nonumber\\
  &\times \prod_{\node\in\leaf{\PT}} p(\rho_\node=0),
\label{eq:treeprior}
\end{align}
where $\U(\cdot)$ denotes the probability mass function of the uniform distribution  over dimensions that contain at least one valid split, and $\U(\cdot |\sdim_{\node})$ denotes the probability density function of the uniform distribution over valid split locations along dimension $\sdim_{\node}$  in block $B_{\node}$.

Given a decision tree $\TS$, the parameters associated with its leaves are independent and identically distributed normal random variables, and so 
\[
p(\bparam|\TS)=\prod_{\node\in\leaf{\T}}\Normal(\mu_{\node}|m_\mu,\sigma_\mu^2). \label{eq:prior:param}
\]
The mean $m_\mu$ and variance $\sigma_\mu^2$ hyperparameters are set indirectly:  %
 \citet{Chipman10} shift and rescale the labels $Y$ such that $y_{\min}=-0.5$ and $y_{\max}=0.5$, and set $m_\mu=0$ and $\sigma_\mu=0.5/k\sqrt{m}$, where $k>0$ is an hyperparameter. This adjustment has the effect of keeping individual node parameters $\mu_\node$ small; the higher the values of $k$ and $m$, the greater the shrinkage towards the mean $m_\mu$. %
 
The prior $p(\sigma^2)$ over the noise variance is an inverse gamma distribution. The hyperparameters $\nu$ and $q$ indirectly control the shape and rate of the inverse gamma prior over $\sigma^2$. \citet{Chipman10} compute an overestimate of the noise variance $\widehat{\sigma}^2$, e.g., using the least-squares variance or the unconditional variance of $Y$, and, for a given shape parameter $\nu$, set the rate such that $\Pr(\sigma\leq\widehat{\sigma})=q$, i.e., the $q$th quantile of the prior over $\sigma$ is located at $\widehat{\sigma}$. 

\citet{Chipman10} recommend the default values: $\nu=3, q=0.9, k=2, \mbart=200$ and $\alpha_s=0.95, \beta_s=2.0$.  Unless otherwise specified, we use this default hyperparameter setting in our experiments.

\subsection{Sequential generative process for trees}\label{sec:seqtreeprior}
\begin{figure*}%
\begin{center}
\subfigure[$\TS_{(0)}$: $E_{(0)}=\{\epsilon\}$] 
{
\hspace{0.12in}
\includegraphics[width=0.10\textwidth]{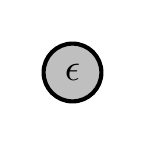}
\hspace{0.175in}
}
\subfigure[$\TS_{(1)}$: $E_{(1)}=\{0,1\}$]{
\includegraphics[width=0.165\textwidth]{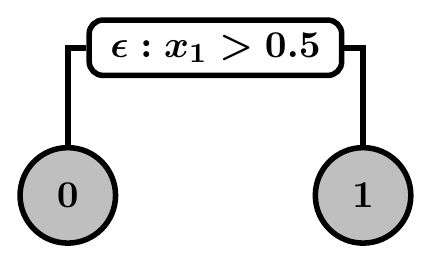}
}
\subfigure[$\TS_{(2)}$: $E_{(2)}=\{1\}$]{
\includegraphics[width=0.165\textwidth]{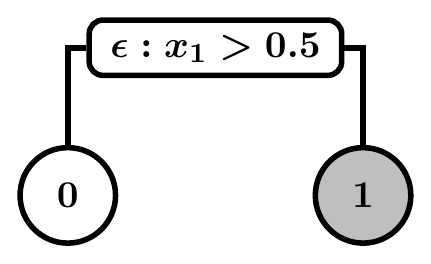}
} 
\subfigure[$\TS_{(3)}$: $E_{(3)}=\{10,11\}$]{
\includegraphics[width=0.21\textwidth]{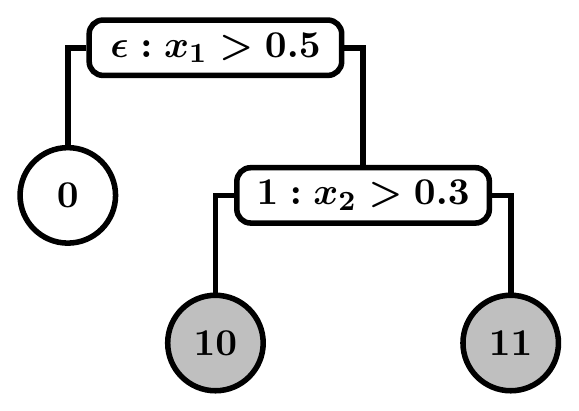}
}
\subfigure[$\TS_{(6)}$: $E_{(6)}=\{\}$]{
\includegraphics[width=0.21\textwidth]{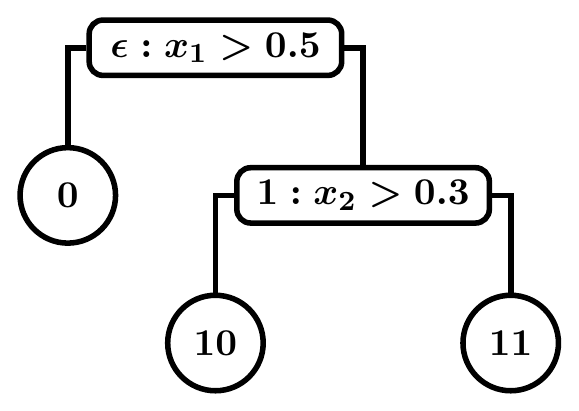}
}
\caption{Sequential generative process for decision trees: 
Nodes \emph{eligible  for expansion} are denoted by the ordered set $E$ and shaded in gray. At iteration 0, we start with the empty tree and $E=\{\epsilon\}$. 
At iteration 1, we pop $\epsilon$ from $E$ and assign it to be an internal node with split dimension $\sdim_{\epsilon}=1$ and split location $\sloc_{\epsilon}=0.5$ and append the child nodes $0$ and $1$ to $E$. At iteration 2, we pop $0$ from $E$ and set it to a leaf node. At iteration 3, we pop $1$ from $E$ and set it to an internal node,  split dimension $\sdim_{1}=2$ and threshold $\sloc_1=0.3$ and append the child nodes $10$ and $11$ to $E$. At iterations $4$ and $5$ (not shown), we pop nodes $10$ and $11$ respectively and assign them to be leaf nodes. At iteration $6$, $E=\{\}$ and the process terminates. By arranging the random variables $\rho$ and $\sdim, \sloc$ (if applicable) for each node in the order of expansion, the tree can be encoded as a sequence.} 
\label{fig:sequentialtree}
\end{center}
\reducevspace
\end{figure*}

\citet{LRT2013} present a sequential generative process for the tree prior $p(\TS|\bX)$, where a tree $\TS$ is generated by starting from an empty tree  $\TS_{(0)}$ and sampling a sequence $\TS_{(1)}, \TS_{(2)}, \dotsc$ of partial trees.\footnote{Note that $\TS_{(t)}$ denotes partial tree at stage $t$, whereas $\TS_j$ denotes the $j$th tree in the ensemble.}
This sequential representation is used as the scaffolding for their SMC algorithm. 
Due to space limitations, we can only briefly review the sequential process. The interested reader should refer to the paper by \citet{LRT2013}:
Let $\TS_{\smciter}=\PT_{\smciter},\bsdim_{\smciter},\bsloc_{\smciter}, E_{\smciter}$ denote the partial tree at stage $\smcitervar$, where $E_{\smciter}$ denotes the ordered set containing the list of nodes \emph{eligible  for expansion} at stage $\smcitervar$. At stage $\smcitervar$, the generative process samples $\TS_{\smciter}$ from $\Pr_{\smcitervar}(\cdot \given \TS_{\smciterold})$ as follows: the first element\protect\footnotemark \ of $E$, say $\node$, is popped and is stochastically assigned to be an internal node or a leaf node 
with probability given by \eqref{eq:splitprob}.
If $\node$ is chosen to be an internal node, we sample the split dimension $\sdim_{\node}$ and split location $\sloc_{\node}$ uniformly among the valid splits, and append $\node0$ and $\node1$ to $E$. Thus, the tree is  expanded in a breadth-wise  fashion and each node is visited just  once. The process terminates when $E$ is empty.  Figure~\ref{fig:sequentialtree} presents a cartoon of the sequential generative process.

\footnotetext{\citet{LRT2013} discuss a more general version where more than one node may be expanded in an iteration. We restrict our attention in this paper to \emph{node-wise expansion}: one node is expanded per iteration and the nodes are expanded in a breadth-wise fashion.}

\begin{algorithm*}%
    \caption{Bayesian backfitting MCMC for posterior inference in BART}
    \label{alg:bart}
    \begin{algorithmic}[1]
    \State Inputs: Training data $(\bX, Y)$, BART hyperparameters $(\nu, q, k, \mbart, \alpha_s, \beta_s)$
    \State Initialization: 
    For all $j$, set $\TS_j^{(0)}= \{ \T_{j}^{(0)} = \{\epsilon\}$, $\bsloc_{j}^{(0)}=\bsdim_{j}^{(0)}=\emptyset\}$ and sample $\bparam_{j}^{(0)}$ 
    \For{$i = 1:\mathsf{max\_iter}$}
    \State Sample $\sigma^{2\ii}| \TS_{\allj}^{\iiold}, \bparam_{\allj}^{\iiold}$ \algcomment{sample from inverse gamma distribution}
    \For{$j=1:\mbart$}
    \State Compute residual $R_{j}^{\ii}$ \algcomment{using (\ref{eq:residual})}

    \State Sample $\TS_{j}^{\ii} | R_{j}^{\ii}, \sigma^{2\ii}, \TS_{j}^{\iiold}$ \algcomment{using \cgm, \growprune\ or PG}
    \State Sample $\bparam_{j}^{\ii} | R_{j}^{\ii}, \sigma^{2\ii}, \TS_{j}^{\ii}$ \algcomment{sample from Gaussian distribution}%
        \EndFor
    \EndFor
\end{algorithmic}
\end{algorithm*} %

\section{Posterior inference for BART}\label{sec:inference}
In this section, we briefly review the MCMC framework proposed in \citep{Chipman10}, discuss limitations of existing samplers  and then present our PG sampler.

\subsection{MCMC for BART}\label{sec:mh}
Given the likelihood and the prior, our goal is to compute the posterior distribution %
\begin{align} %
p(\ensemble,\sigma^2|Y,\bX) \ \propto \qquad\qquad\qquad\qquad\quad \nonumber\\
\quad \ell(Y|\ensemble,\sigma^2,\bX) \ p(\ensemble,\sigma^2|\bX). \label{eq:posterior}
\end{align}
 \citet{Chipman10} proposed a Bayesian backfitting MCMC to sample from the BART posterior. At a high level, the Bayesian backfitting MCMC is a Gibbs sampler that loops through the trees, sampling each tree $\TS_j$ and associated parameters  $\bparam_j$  conditioned on $\sigma^2$ and the remaining trees and their associated parameters $\ensemblenotj$, and samples $\sigma^2$ conditioned on all the trees and parameters $\ensemble$. 
Let $\TS_{j}^{\ii},\param_{j}^{\ii},$ and $\sigma^{2\ii}$ respectively denote the values of $\TS_{j}, \param_{j}$ and $\sigma^2$ at the $i^{th}$ MCMC iteration. 
Sampling $\sigma^2$ conditioned on $\ensemble$ is straightforward due to conjugacy. 
To sample 
$\TS_j, \bparam_j$ conditioned on the other trees $\ensemblenotj$,  we first sample 
$\TS_j|\ensemblenotj,\sigma^2$ and then sample $\bparam_j|\TS_j,\ensemblenotj,\sigma^2$. (Note that $\bparam_j$ is integrated out while sampling $\TS_j$.)  
More precisely, we compute the residual 
\begin{align} 
R_{j}=& \ Y - \textstyle \sum_{\jdash=1,\jdash\neq j}^{m} g(\bX; \TS_{\jdash}, \param_{\jdash}). 
\label{eq:residual}
\end{align}
Using the residual $R_{j}^{\ii}$ as the target, \citet{Chipman10} sample $\TS_j^{\ii}$ by proposing local changes to $\TS_j^{\iiold}$. 
Finally,  $\bparam_j$ is sampled from a Gaussian distribution conditioned on $\TS_j,\ensemblenotj,\sigma^2$. The procedure is summarized in Algorithm~\ref{alg:bart}.

\subsection{Existing samplers for BART}\label{sec:existing samplers}
To sample $\TS_j$, \citet{Chipman10} use the MCMC algorithm proposed by \citet{chipman1998bayesian}.  This algorithm, which we refer to as \cgm. \cgm\ is a \emph{Metropolis-within-Gibbs} sampler that randomly  chooses one of the following four moves: \grow\ (which randomly chooses a leaf node and splits it further into left and right children), \prune\ (which randomly chooses an internal node where both the children are leaf nodes and prunes the two leaf nodes, thereby making the internal node a leaf node), \change\ (which changes the decision rule at a randomly chosen internal node), \swap\ (which swaps the decision rules at a parent-child pair where both the parent and child are internal nodes). There are two issues with the \cgm\ sampler: 
(1) the \cgm\ sampler makes local changes to the tree, which is known to affect mixing when computing the posterior over a single decision tree  \citep{wu2007bayesian}.  \citet{Chipman10} claim that the default hyper-parameter values encourage shallower trees and hence mixing is not affected significantly. %
However, if one wishes to use BART on large datasets where individual trees are likely to be deeper, the \cgm\ sampler might suffer from mixing issues. %
(2) The \emph{change} and \emph{swap} moves in \cgm\ sampler are computationally expensive for large datasets that involve deep trees (since they involve re-computation %
 of all likelihoods in the subtree below the top-most node affected by the proposal). For computational efficiency, \citet{pratola2013parallel} propose  using only the \emph{grow} and \emph{prune} moves; we will call this the \growprune\  sampler. However, as we illustrate in section \ref{sec:experiments}, the  \growprune\  sampler can inefficiently explore the posterior in scenarios where there are multiple possible trees that explain the observations equally well.
In the next section, we present a novel sampler that addresses both of these concerns.

\reducevspace
\subsection{PG sampler for BART}\label{sec:pg proposal}
Recall that \citet{Chipman10} 
 sample $\TS_j^{\ii}$ using $R_{j}^{\ii}$ as the target by proposing local changes to $\TS_j^{\iiold}$. 
 It is natural to ask if it is possible to sample a complete tree $\TS_j^{\ii}$ rather than just  local changes.  Indeed, this is possible by marrying the sequential representation of the tree proposed by \citet{LRT2013}  (see section \ref{sec:seqtreeprior}) with the Particle Markov Chain Monte Carlo (PMCMC) framework \citep{AndDouHol2010a} where an SMC algorithm (particle filter) is used as a high-dimensional proposal for MCMC. 
 The PG %
sampler is implemented using the so-called \emph{conditional SMC} algorithm 
(instead of the Metropolis-Hastings samplers described in Section~\ref{sec:existing samplers}) 
 in line 7 of Algorithm~\ref{alg:bart}. At a high level, the conditional SMC algorithm is similar to the SMC algorithm proposed by \citet{LRT2013}, except that one of the particles is clamped to the current tree $\TS_{j}^{\iiold}$. 
 
Before describing the  PG sampler, we derive the conditional posterior $\TS_j|\ensemblenotj,\sigma^2,Y,\bX$. 
Let $N(\node)$ denote the set of data point indices $n \in \{1,\dots,N\}$ such that $\bx_n \in B_{\node}$. Slightly abusing the notation, let $R_{N({\node})}$  denote the vector containing residuals of data points in node $\node$. 
Given $R \defas Y - \sum_{\jdash\neq j} g(\bX; \TS_{\jdash}, \param_{\jdash})$, 
it is easy to see that the conditional posterior over $\TS_j,\bparam_j$ is given by
\begin{align*} %
& p(\TS_j,\bparam_j|\ensemblenotj,\sigma^2,Y,\bX) \\
&\propto p(\TS_j|\bX) \hspace{-0.1in}
\prod_{\node\in\leaf{\PT_j}} 
\prod_{n\in N(\node)} \Normal(R_{n}|\mu_\node,\sigma^2)
\Normal(\mu_\node|m_\mu,\sigma_\mu^2).
\end{align*}
Let $\pi(\TS_j)$ denote the conditional posterior over $\TS_j$. Integrating out $\bparam$ and using (\ref{eq:treeprior}) for $p(\TS_j|\bX)$, %
the conditional posterior $\pi(\TS_j)$ is 
\begin{align} %
&\pi(\TS_j) =p(\TS_j|\ensemblenotj,\sigma^2,Y,\bX) \qquad\qquad\qquad\nonumber\\ 
&\qquad\propto 
 p(\TS_j|\bX)
 \prod_{\node\in\leaf{\PT_j}} p(R_{N(\node)}|\sigma^2,m_\mu,\sigma_\mu^2), \label{eq:posterior:tj}
\end{align}
where $p(R_{N(\node)}|\sigma^2,m_\mu,\sigma_\mu^2)$ denotes the marginal likelihood at a node $\node$, given  by
 \begin{align} %
&p(R_{N(\node)}|\sigma^2,m_\mu,\sigma_\mu^2)  \nonumber\\
&= \int_{\mu_\node}
\prod_{n\in N(\node)}\Normal(R_{n}|\mu_\node,\sigma^2)
\Normal(\mu_\node|m_\mu,\sigma_\mu^2)d\mu_\node.
\label{eq:marginal:likelihood:node}
\end{align}

The goal is to sample from the (conditional) posterior distribution $\pi(\TS_j)$. %
 \citet{LRT2013} presented a top-down particle filtering algorithm that approximates the posterior over  decision trees. Since this SMC algorithm can sample complete trees, it is tempting to substitute an exact sample from $\pi(\TS_j)$ with an approximate sample from the particle filter. 
However, 
\citet{AndDouHol2010a} observed that this naive approximation does not leave the joint posterior distribution (\ref{eq:posterior}) invariant, 
and so they proposed instead to generate a sample using a modified version of the SMC algorithm, which they called the \emph{conditional-SMC algorithm}, and demonstrated that this leaves the joint distribution (\ref{eq:posterior}) invariant. 
 (We refer the reader to the paper by \citet{AndDouHol2010a} for further details about the PMCMC framework.)   
By building off the top-down particle filter for decision trees,
we can define a conditional-SMC algorithm for sampling from $\pi(\TS_j)$.

The conditional-SMC algorithm 
is an MH kernel with $\pi(\TS_j)$ as its stationary distribution. %
To reduce clutter, let $\TS^*$ denote the old tree and $\TS$ denote the tree we wish we to sample. %
The conditional-SMC algorithm samples $\TS$ from a $C$-particle approximation of $\pi(\TS)$, which can be written as $\sum_{c=1}^{\nparts} \wtbar\cc \delta_{\TS\cc}$ where $\TS\cc$ denotes the $c^{th}$ tree (particle) and the weights sum to 1, that is, $\sum_c\wtbar\cc=1$. 

\textbf{SMC proposal}: Each particle $\TS\cc$ is the end product of a sequence of partial trees $\TS_{(0)}\cc, \TS_{(1)}\cc, \TS_{(2)}\cc, \dotsc$, and the weight $\wtbar\cc$ reflects how well the $c^{th}$ tree explains the residual $R$. 
One of the particles, say the first particle, without loss of generality, is clamped to the old tree $\TS^*$ at all stages of the particle filter, %
 i.e., $\TS_{\smciter}(1)=\TS_{\smciter}^*$.  At stage $\smcitervar$, the remaining 
 $\nparts-1$ particles are sampled from the sequential generative process 
 $\Pr_{\smcitervar}(\cdot \given \TS_{\smciterold}(c))$
 described in section~\ref{sec:seqtreeprior}.
Unlike state space models where the length of the latent state sequence is fixed, the sampled decision tree sequences may be of different length and could potentially be deeper than the old tree $\TS^*$. Hence, whenever $E_{\smciter}=\emptyset$, we set $\Pr_{\smciter}(\TS_{\smciter}|\TS_{\smciterold})=\delta_{\TS_{\smciterold}}$, i.e., $\TS_{\smciter}=\TS_{\smciterold}$.

\textbf{SMC weight update}: Since the prior is used as the proposal, the particle weight $\wt_{\smciter}(c)$ is multiplicatively updated with the ratio of the marginal likelihood of $\TS_{\smciter}(c)$ to the  marginal likelihood of $\TS_{\smciterold}(c)$.  The marginal likelihood associated with a (partial) tree $\TS$ is a product of the marginal likelihoods associated with the leaf nodes of $\TS$ defined in (\ref{eq:marginal:likelihood:node}). Like \citet{LRT2013}, we treat the eligible nodes $E_{\smciter}$ as leaf nodes while computing the marginal likelihood for a partial tree $\TS_{\smciter}$. Plugging in (\ref{eq:marginal:likelihood:node}), the SMC weight update is given by (\ref{eq:smcweight}) in Algorithm~\ref{alg:smc}. 

\textbf{Resampling}: The resampling step in the conditional-SMC algorithm is slightly different from the typical SMC resampling step. Recall that the first particle is always clamped to the old tree. The remaining $\nparts-1$ particles are resampled such that the probability  of choosing particle $c$ is proportional to its weight $\wt_{\smciter}(c)$. We used multinomial resampling in our experiments, although other resampling strategies are possible. 

When none of the trees contain eligible nodes, the conditional-SMC algorithm stops and returns a sample from the particle approximation. Without loss of generality, we assume that the $C^{th}$ particle is returned. The  PG sampler is summarized in Algorithm~\ref{alg:smc}. 

The computational complexity of the conditional-SMC algorithm in Algorithm~\ref{alg:smc} 
is similar to that of the top-down algorithm~\citep[][\S3.2]{LRT2013}. 
Even though the  PG sampler has a higher per-iteration complexity in general compared to \growprune\ and \cgm\ samplers, it can mix faster since it can propose a completely different tree that explains the data. The \growprune\ sampler requires many iterations to explore multiple modes (since a prune operation is likely to be rejected around a mode). 
The \cgm\ sampler can change the decisions at internal nodes; however, it is inefficient since a change in an internal node that leaves any of the nodes in the subtree below empty will be rejected. 
We demonstrate the competitive  performance of PG in the experimental section.

\begin{algorithm*}%
  \caption{Conditional-SMC algorithm used in the PG-BART sampler}   
\label{alg:smc}                       
\begin{algorithmic}[1]   
    \State Inputs: Training data: features $\bX$, `target' $R$ \algcomment{$R$ denotes residual in BART}\\
                             \hspace{12.5mm} Number of particles $\nparts$ \\
                             \hspace{12.5mm} Old tree $\TS^*$ (along with the partial tree sequence $\TS^*_{(0)},\TS^*_{(1)},\TS^*_{(2)},\dotsc$)
    \State Initialize: For $\Cc=1:\nparts$,  set  
    $\T_{(0)}\cc = E_{(0)}\cc = \{\epsilon\}$ and $\bsloc_{(0)}\cc=\bsdim_{(0)}\cc=\emptyset$ \\
  \hspace{15mm}    For $\Cc=1:\nparts$,  set weights $\wt_{(0)}\cc = p(R_{N(\epsilon)}|\sigma^2,m_\mu,\sigma_\mu^2)$  
  and $W_{(0)} = \sum_{\Cc} \wt_{(0)}\cc$
 \For{$\smcitervar = 1:$ \textsf{max-stages}} 
         \State Set $\TS_{\smciter}\ccone = \TS_{\smciter}^*$ \algcomment{clamp the first particle to the partial tree of $\TS^*$ at stage $\smcitervar$}

    \For{$\Cc = 2:\nparts$}
    \State \hspace{-0.1in}Sample $\TS_{\smciter}\cc$ from $\Pr_{\smciter}(\cdot \given \TS_{\smciterold}\cc)$
 where $\TS_{\smciter}\cc \defas (\T_{\smciter}\cc,\bsdim_{\smciter}\cc,\bsloc_{\smciter}\cc,E_{\smciter}\cc)$ \algcomment{section \ref{sec:seqtreeprior}}
 \EndFor
 
  \For{$\Cc = 1:\nparts$}
   \State  \LineComment{If $E_{\smciterold}(c)$ is non-empty, let $\node$ denote the node popped from $E_{\smciterold}(c)$.}  
   \State Update weights:$\vphantom{\TS^\cc}$ 
\begin{numcases}{\wt_{\smciter}\cc =  \hspace{-0.05in}}
\wt_{\smciterold}\cc  & \hspace{-0.2in} \textit{if} $ E_{\smciterold}(c)$ \textit{is empty or $\node$ is stopped}, \nonumber\\
\wt_{\smciterold}\cc \dfrac { \prod_{\node' = \node0,\node1} p(R_{N(\node')}|\sigma^2,m_\mu,\sigma_\mu^2) }
                { p(R_{N(\node)}|\sigma^2,m_\mu,\sigma_\mu^2)} 
                      & \hspace{-0.2in} \textit{if $\node$ is split. } \label{eq:smcweight}
\end{numcases}

    \EndFor
    \State Compute normalization: 
    $W_{\smciter} = \sum_{\Cc} \wt_{\smciter}\cc$
    \State Normalize weights: $(\forall \Cc)\, \wtbar_{\smciter}\cc = \wt_{\smciter}\cc / W_{\smciter}$
 \State Set $j_1=1$ and for $\Cc=2:\nparts$, 
    resample indices $j_\Cc$ from $\sum_{\Ccd} \wtbar_{\smciter}\ccd \delta_{\Ccd}$  \algcomment{resample all particles except the first}
    \State $(\forall \Cc)$ $\TS_{\smciter}\cc \gets \TS_{\smciter}{(j_\Cc)}$; $\wt_{\smciter}\cc\gets W_{\smciter}/\nparts$
 
    \If{$(\forall \Cc)\, E_{\smciter}\cc = \emptyset$}
         exit for loop
    \EndIf
    \EndFor
\Return $\TS_{\smciter}(\nparts)=(\PT_{\smciter}(\nparts),\bsdim_{\smciter}(\nparts), \bsloc_{\smciter}(\nparts))$ \algcomment{return a sample from     the  approximation  $\sum_{\Ccd} \wtbar_{\smciter}\ccd \delta_{\TS_{\smciter}\ccd}$ to  line 7 of Algorithm~\ref{alg:bart}}
\end{algorithmic}
\end{algorithm*} %

\section{Experimental evaluation}\label{sec:experiments}
In this section, we present experimental comparisons between the  PG sampler and existing samplers for BART. Since the main contribution of this work is a different inference algorithm for an existing model, we just compare the efficiency of the inference algorithms and do not compare to other models. BART has been shown to demonstrate excellent prediction performance compared to other popular black-box non-linear regression approaches; we refer the interested reader to \citet{Chipman10}.

We implemented all the samplers in Python and ran experiments on the same desktop machine so that the timing results are comparable. 
The scripts can be downloaded from the authors' webpages.  %
We set the number of particles $C=10$ for computational efficiency and $\textsf{max-stages}=5000$, following \citet{LRT2013}, although the algorithm always terminated much earlier.

\subsection{Hypercube-$D$ dataset}
We investigate the performance of the samplers on a dataset where there are multiple trees that explain the residual (conditioned on other trees). This problem is equivalent to posterior inference over a decision tree where the labels are equal to the residual. %
 Hence, we generate a synthetic dataset where multiple trees are consistent with the observed labels.  Intuitively, a local sampler can be expected to mix reasonably well when the true posterior consists of shallow trees; however, a local sampler will lead to an inefficient exploration when the posterior consists of deep trees. Since the depth of trees in the true posterior is at the heart of the mixing issue, we create synthetic datasets where the depth of trees in the true posterior can be controlled.%

\begin{figure}[h!]%
\begin{center}
\includegraphics[width=0.828\columnwidth]{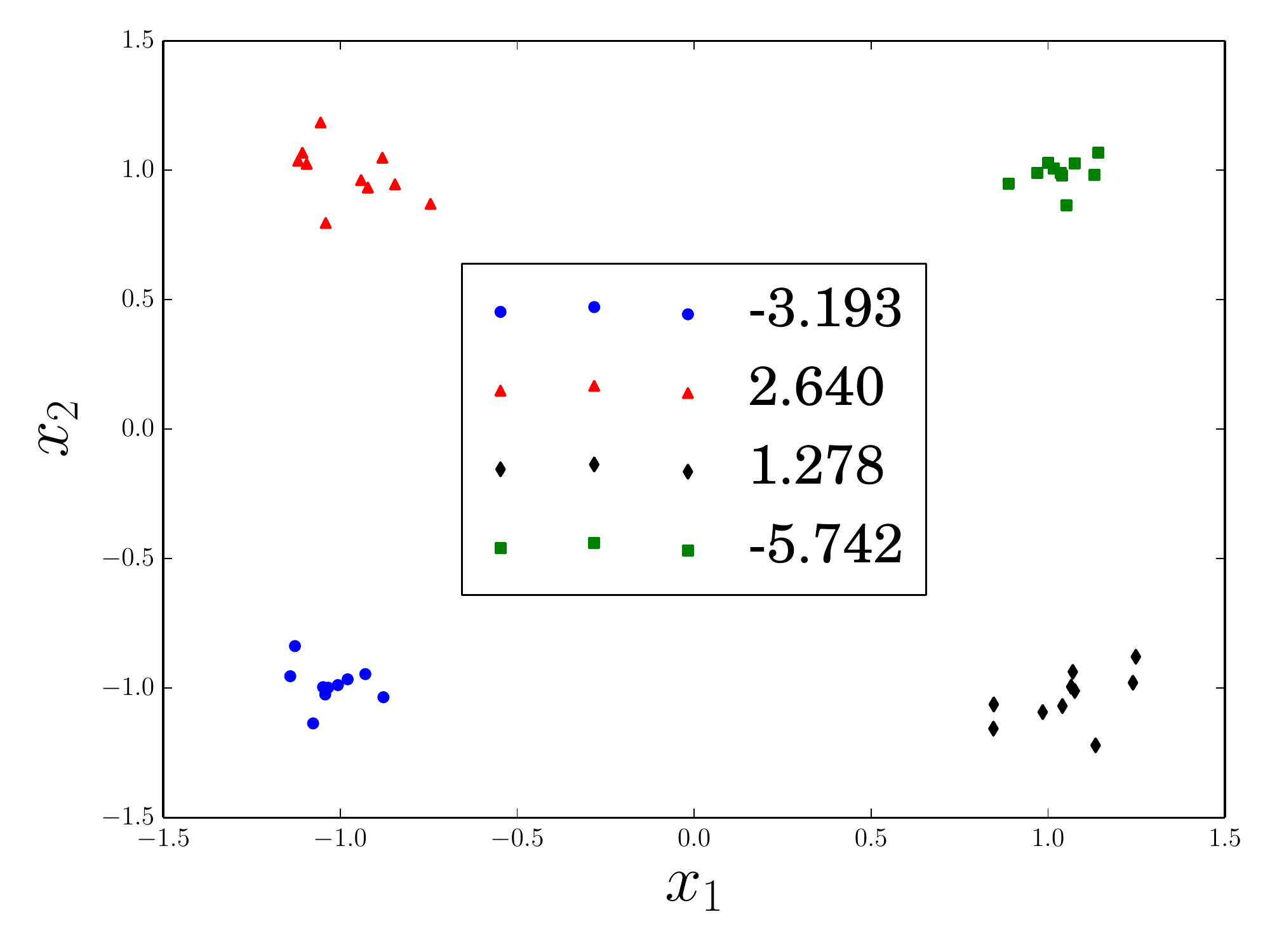} %
\caption{Hypercube-2 dataset: 
see main text for details.}%
\label{fig:hypercube-2:data}
\end{center}
\reducevspace
\reducevspace
\end{figure}

We generate the hypercube-$D$ dataset as follows: for each of the $2^D$ vertices of $[-1,1]^D$, we sample 10 data points. The $\bx$ location of a data point is generated as $\bx=\bv+\epsilon$ where $\bv$ is the vertex location and $\epsilon$ is a random offset generated as $\epsilon\sim\Normal(\bm{0},0.1^2I_D)$. Each vertex is associated with a different function value and the function values are generated from $\Normal(0, 3^2)$. Finally the observed label is generated as $y=f+e$ where $f$ denotes the true function value at the vertex and $e\sim\Normal(0,0.01^2)$. 
Figure~\ref{fig:hypercube-2:data} shows a sample hypercube-$2$ dataset. %
As $D$ increases, the number of trees that explains the observations increases. %

We fix $\mbart=1, \alpha_s=0.95$ and set remaining BART hyperparameters to the default values. Since the true tree has $2^D$ leaves, we set\footnote{The values of $\beta_s$ for $D=2,3,4,5$ and $7$ are $1.0,  0.5, 0.4, 0.3$ and  $0.25$ respectively.} $\beta_s$ such that the expected number of leaves is roughly $2^D$. We run  2000 iterations of MCMC. %
  Figure~\ref{fig:hypercube-4} 
 illustrates the posterior trace plots for $D=4$. (See supplemental material for additional posterior trace plots.)  
 We observe that PG converges much faster to the posterior in terms of number of leaves as well as the test MSE. %
  We observe that \growprune\ sampler tends to overestimate the number of leaves; the low value of train MSE indicates that the \growprune\ sampler is stuck close to a mode and is unable to explore the true posterior.  \citet{pratola2013efficient}  has reported similar behavior  of \growprune\  sampler on a different dataset as well. 

\begin{figure}[h!]%
\begin{center}
\includegraphics[width=0.99\columnwidth]{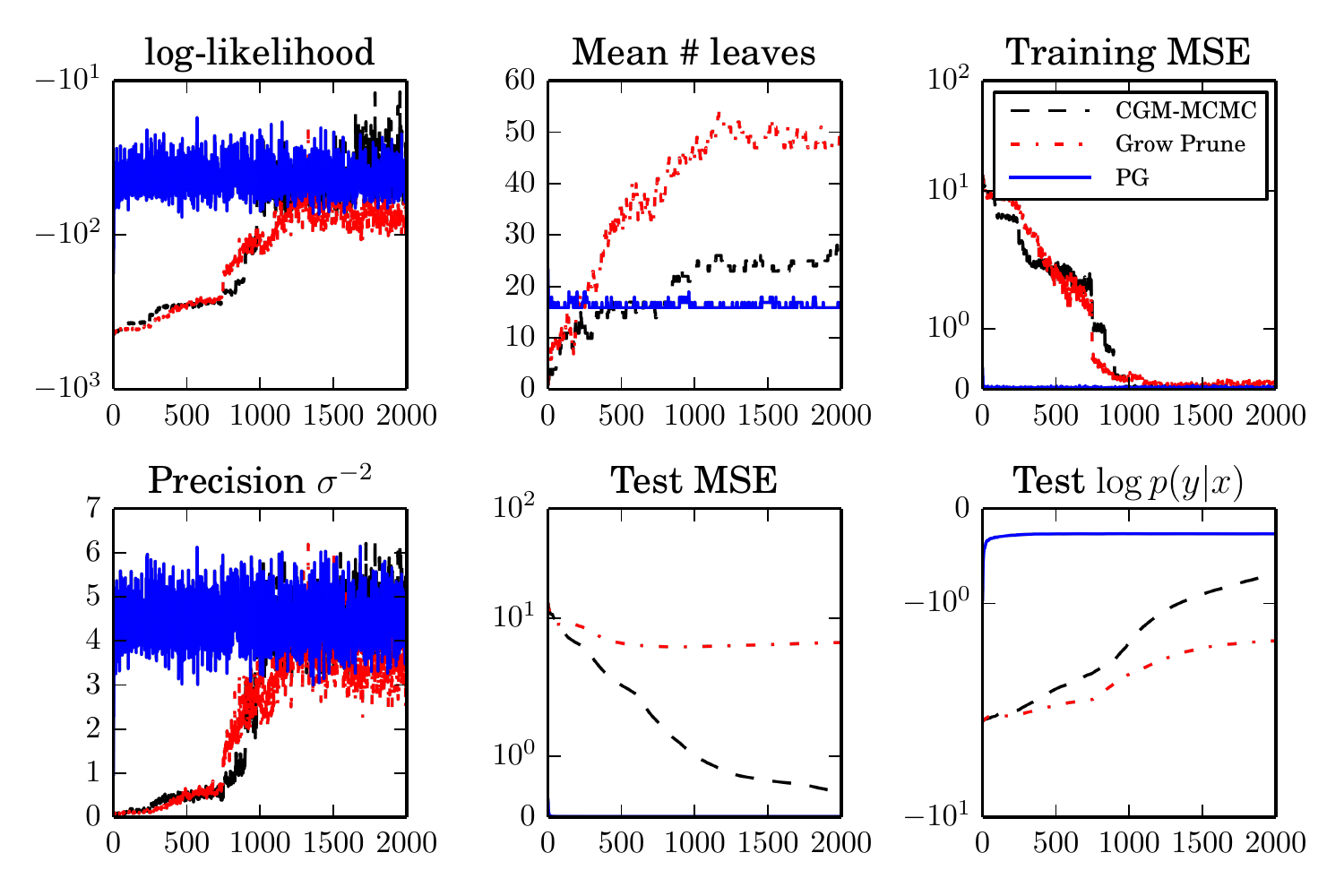} %
\caption{Results on Hypercube-4 dataset.}
\label{fig:hypercube-4}
\end{center}
\end{figure}

We compare the  algorithms by computing effective sample size (ESS). ESS is a measure of how well the chain mixes and is frequently used to assess performance of MCMC algorithms; we compute ESS using R-CODA \citep{plummer2006coda}. We discard the first 1000 iterations as burn-in and use the remaining 1000 iterations to compute ESS. Since the per iteration cost of generating a sample differs across samplers, we additionally report ESS per unit time. 
The ESS (computed using log-likelihood values) and ESS per second (ESS/s) values are shown in Tables~\ref{tab:ess:hypercube} and \ref{tab:esspertime:hypercube} respectively. When the true tree is shallow ($D=2$ and $D=3$), we observe that \cgm\ sampler mixes well and is computationally efficient. However, as the depth of the true tree increases ($D=4,5,7$), PG achieves much higher ESS and ESS/s compared to \cgm\ and \growprune\ samplers.
\begin{table}%
\begin{center}
\begin{tabular}{|c|c|c|c|}
\hline
$D$&\cgm\   & \growprune\   & PG  \\ %
\hline
2&\textbf{751.66}&473.57&259.11\\ 
3&\textbf{762.96}&285.2&666.71 \\ 
4&14.01&11.76&\textbf{686.79} \\ 
5&2.92&1.35&\textbf{667.27} \\ 
7&1.16&1.78&\textbf{422.96}\\ 
\hline
\end{tabular}
\end{center}
\caption{Comparison of ESS for \cgm, \growprune\ and  PG samplers on Hypercube-$D$ dataset.}
\label{tab:ess:hypercube}
\end{table}
\begin{table}%
\begin{center}
\begin{tabular}{|c|c|c|c|}
  \hline
$D$& \cgm & \growprune   & PG   \\
\hline
2& \textbf{157.67}& 114.81 & 7.69\\ 
3& \textbf{93.01}& 26.94 & 11.025\\ 
4& 0.961 & 0.569 &\textbf{5.394}\\ 
5& 0.130 & 0.071 &\textbf{1.673}\\ 
7& 0.027 & 0.039 &\textbf{0.273}\\ 
\hline
\end{tabular}
\end{center}
\caption{Comparison of ESS/s (ESS per second) for \cgm, \growprune\ and  PG samplers on Hypercube-$D$ dataset.}
\label{tab:esspertime:hypercube}
\end{table}

\subsection{Real world datasets}
In this experiment,  we study the effect of the data dimensionality on mixing. Even when the trees are shallow, 
the number of trees consistent with the labels increases as the data dimensionality increases. Using the default BART prior (which promotes shallower trees), we compare the performance of the samplers on real world datasets of varying dimensionality. 

We consider the  \houses, \msd\ and \ctslices\ datasets used by \citet{johnson2011learning}. For each dataset, there are three training sets, each of which contains 2000 data points, and a single test set. The dataset characteristics are summarized in Table~\ref{tab:datasets}. %
\begin{table}[h!]%
\begin{center}
\begin{tabular}{|c|c|c|c|}
\hline
Dataset & $N_{\mathsf{train}}$ & $N_{\mathsf{test}}$ & D \\ \hline
\houses & 2000 & 5000 & 6 \\  
\msd & 2000 & 51630 & 90 \\  
\ctslices & 2000 & 24564 & 384 \\ 
\hline
\end{tabular}
\end{center}
\caption{Characteristics of datasets.}
\label{tab:datasets}
\end{table}

We run each sampler using the three training datasets and report average ESS and ESS/s. All three samplers achieve very similar MSE to those reported by \citet{johnson2011learning}. The average number of leaves in the posterior trees was found to be small and
very similar for all the samplers. %
Tables~\ref{tab:ess:realworld} and \ref{tab:esspertime:realworld} respectively present results comparing ESS and ESS/s of the different samplers. As the %
data dimensionality  increases, we observe that PG outperforms existing samplers.

\begin{table}[h!]%
\begin{center}
\resizebox{\linewidth}{!}{ %
\begin{tabular}{|c|c|c|c|}
\hline
$Dataset$&\cgm & \growprune & PG   \\ \hline
\houses & 18.956 & 34.849 & \textbf{76.819} \\
\msd  & 29.215 & 21.656 & \textbf{76.766} \\
\ctslices & 2.511 & 5.025 & \textbf{11.838} \\
\hline
\end{tabular}
} %
\end{center}
\caption{Comparison of ESS for \cgm, \growprune\ and  PG samplers on real world datasets.}
\label{tab:ess:realworld}
\end{table}

\begin{table}[h!]%
\begin{center}
\resizebox{1.0\linewidth}{!}{ %
\begin{tabular}{|c|c|c|c|}
\hline
$Dataset$&\cgm  & \growprune  & PG  \\ 
& $\times10^{-3}$ &  $\times10^{-3}$ &    $\times10^{-3}$  \\  
\hline
\houses &  1.967 & \textbf{48.799} & 16.743 \\
\msd & 2.018 & 7.029 & \textbf{ 14.070} \\
\ctslices & 0.080 &  0.615 & \textbf{2.115} \\
\hline
\end{tabular}
} %
\end{center}
\caption{Comparison of ESS/s for \cgm, \growprune\ and  PG samplers on real world datasets.}
\label{tab:esspertime:realworld}
\end{table}

\section{Discussion}%
We have presented a novel  PG sampler for BART. Unlike existing samplers which make local moves, PG can propose complete trees. Experimental results confirm that PG dramatically increases mixing %
when the true posterior consists of  deep trees or when the data dimensionality is high. While we have presented PG only for the BART model, it is applicable to extensions of BART that use a different likelihood model as well. %
 PG can also be used along with other priors for decision trees, e.g., those of~\citet{denison1998bayesian}, \citet{wu2007bayesian} and \citet{LRT2014}. 
{Backward simulation \citep{lindsten2013backward} and ancestral sampling \citep{lindsten2012ancestor} have been shown to significantly improve mixing of PG for state-space models. Extending these ideas to PG-BART is a challenging and interesting future direction.}%

\section*{Acknowledgments} %
BL gratefully acknowledges generous funding from the Gatsby Charitable Foundation. 
This research was carried out in part while DMR held a Research Fellowship at Emmanuel College, Cambridge, with funding also from a Newton International Fellowship through the Royal Society. 
YWT's research leading to these results has received funding from EPSRC (grant
EP/K009362/1) and the ERC under the EU's FP7 Programme (grant agreement no. 617411).

\newpage %
\bibliography{bart-pmcmc}
\bibliographystyle{abbrvnat}

\clearpage
\newpage
{\center \Large{\textbf{Supplementary material}}}
\appendix

\section{Results on hypercube$-D$ dataset}
\begin{figure}[h!]%
\begin{center}
\includegraphics[width=0.99\columnwidth]{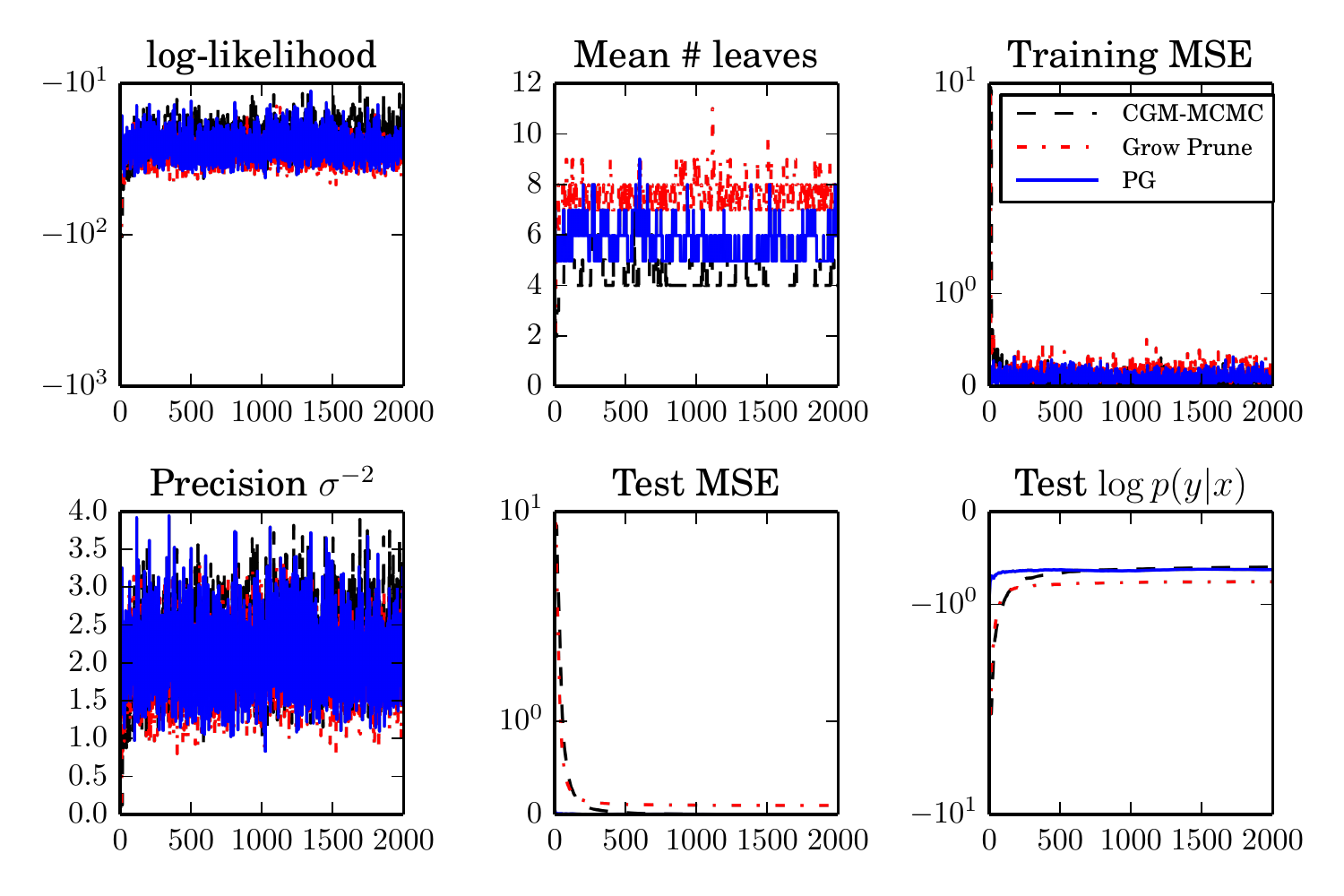} %
\caption{Results on Hypercube-2 dataset.}
\label{fig:hypercube-2}
\end{center}
\end{figure}

\begin{figure}[h!]%
\begin{center}
\includegraphics[width=0.99\columnwidth]{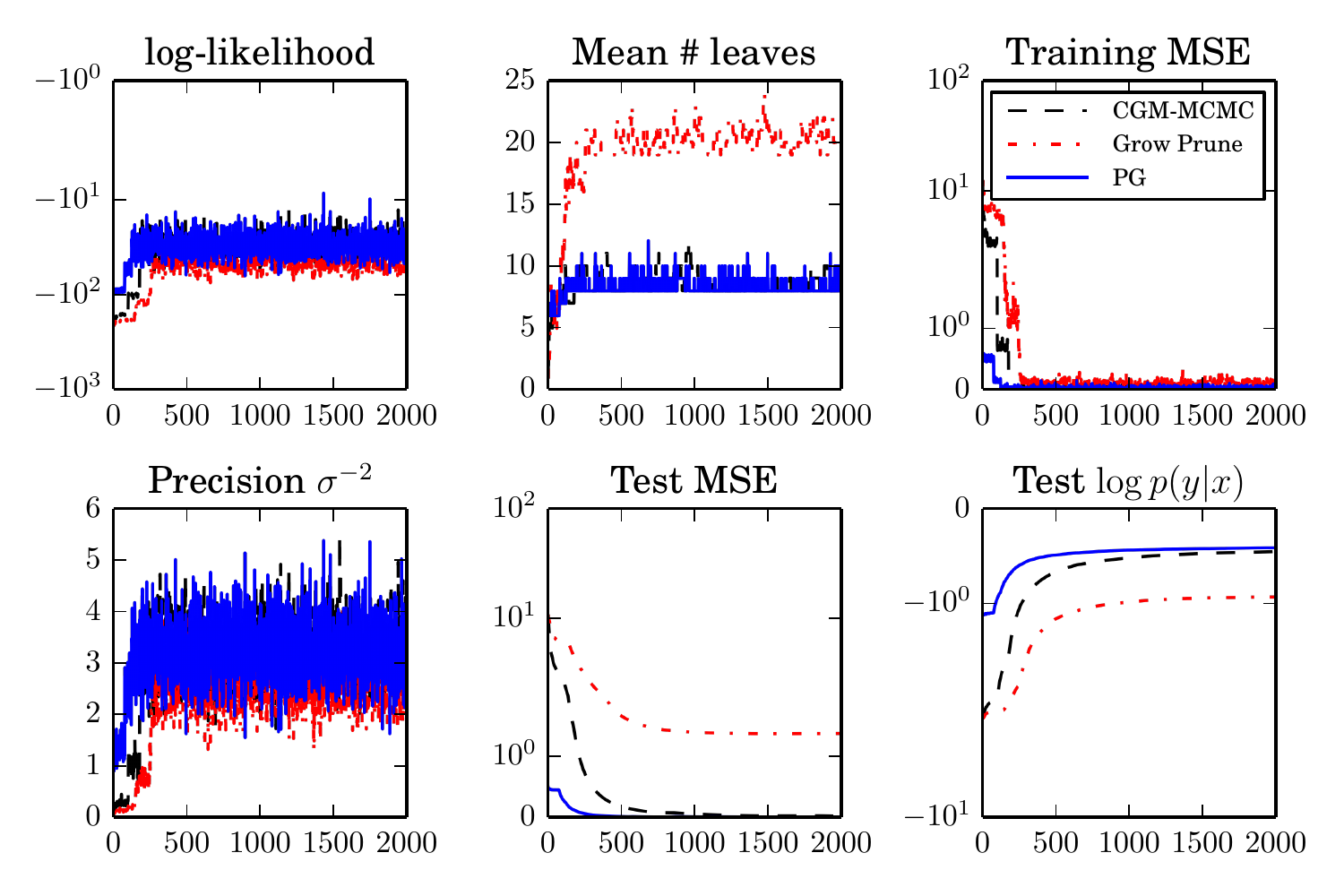} %
\caption{Results on Hypercube-3 dataset.}
\label{fig:hypercube-3}
\end{center}
\end{figure}

\end{document}